\documentclass[sigconf, screen]{acmart}
\acmSubmissionID{6312}
\renewcommand\footnotetextcopyrightpermission[1]{} 
\usepackage{lineno,hyperref}
\usepackage{bbm}
\usepackage{bm}
\usepackage{amsfonts}
\usepackage{amsmath}
\usepackage{mathrsfs}
\usepackage{graphicx}
\usepackage{caption}
\usepackage{subfigure}
\usepackage{booktabs}
\usepackage{setspace}
\usepackage{tabularx}
\usepackage{makecell}
\usepackage{multirow}
\usepackage{bbding}
\usepackage{soul}
\usepackage{euscript}
\AtBeginDocument{%
	\providecommand\BibTeX{{%
			\normalfont B\kern-0.5em{\scshape i\kern-0.25em b}\kern-0.8em\TeX}}}

\begin{document}

\title{Dual-Granularity Cross-Modal Identity Association for Weakly-Supervised Text-to-Person Image Matching}


%
%
%
%
%


\author{Yafei Zhang}
\affiliation{%
	\institution{Kunming University of Science and Technology}
	\city{Kunming}
	\country{China}
	}
\email{zyfeimail@163.com}

\author{Yongle Shang}
\affiliation{%
	\institution{Kunming University of Science and Technology}
	\city{Kunming}
	\country{China}
	}
\email{sylmail99@163.com}

\author{Huafeng Li}
\affiliation{%
	\institution{Kunming University of Science and Technology}
	\city{Kunming}
	\country{China}
	}
\email{hfchina99@163.com}
\authornote{Corresponding author.}



\begin{abstract}
Weakly supervised text-to-person image matching, as a crucial approach to reducing models' reliance on large-scale manually labeled samples, holds significant research value. However, existing methods struggle to predict complex one-to-many identity relationships, severely limiting performance improvements. To address this challenge, we propose a local-and-global dual-granularity identity association mechanism. Specifically, at the local level, we explicitly establish cross-modal identity relationships within a batch, reinforcing identity constraints across different modalities and enabling the model to better capture subtle differences and correlations. At the global level, we construct a dynamic cross-modal identity association network with the visual modality as the anchor and introduce a confidence-based dynamic adjustment mechanism, effectively enhancing the model’s ability to identify weakly associated samples while improving overall sensitivity.
Additionally, we propose an information-asymmetric sample pair construction method combined with consistency learning to tackle hard sample mining and enhance model robustness. Experimental results demonstrate that the proposed method substantially boosts cross-modal matching accuracy, providing an efficient and practical solution for text-to-person image matching.
\end{abstract}



\maketitle

\section{Introduction}
Text-to-person image matching \cite{32} (also known as Text-based person search or Text-to-image person re-identification) is emerging as a key technology in applications such as intelligent security and smart cities. It aims to perform cross-modal matching between natural language descriptions of pedestrians and large-scale image databases, enabling accurate retrieval of target individuals without the need for visual queries. This technology holds strong potential in public security scenarios, including criminal investigations, missing person searches, and intelligent video surveillance. Mainstream supervised methods for text-to-person image matching rely on large-scale, high-quality labeled image-text datasets  \cite{14,15,32,35,36,37,43} and employ complex deep neural networks for representation alignment \cite{2,4,5,6,46,47,31}. However, the high cost of annotation presents a major challenge, significantly limiting their scalability in real-world applications.

\begin{figure}[t!]
	\centering
	\includegraphics[width=0.8\linewidth]{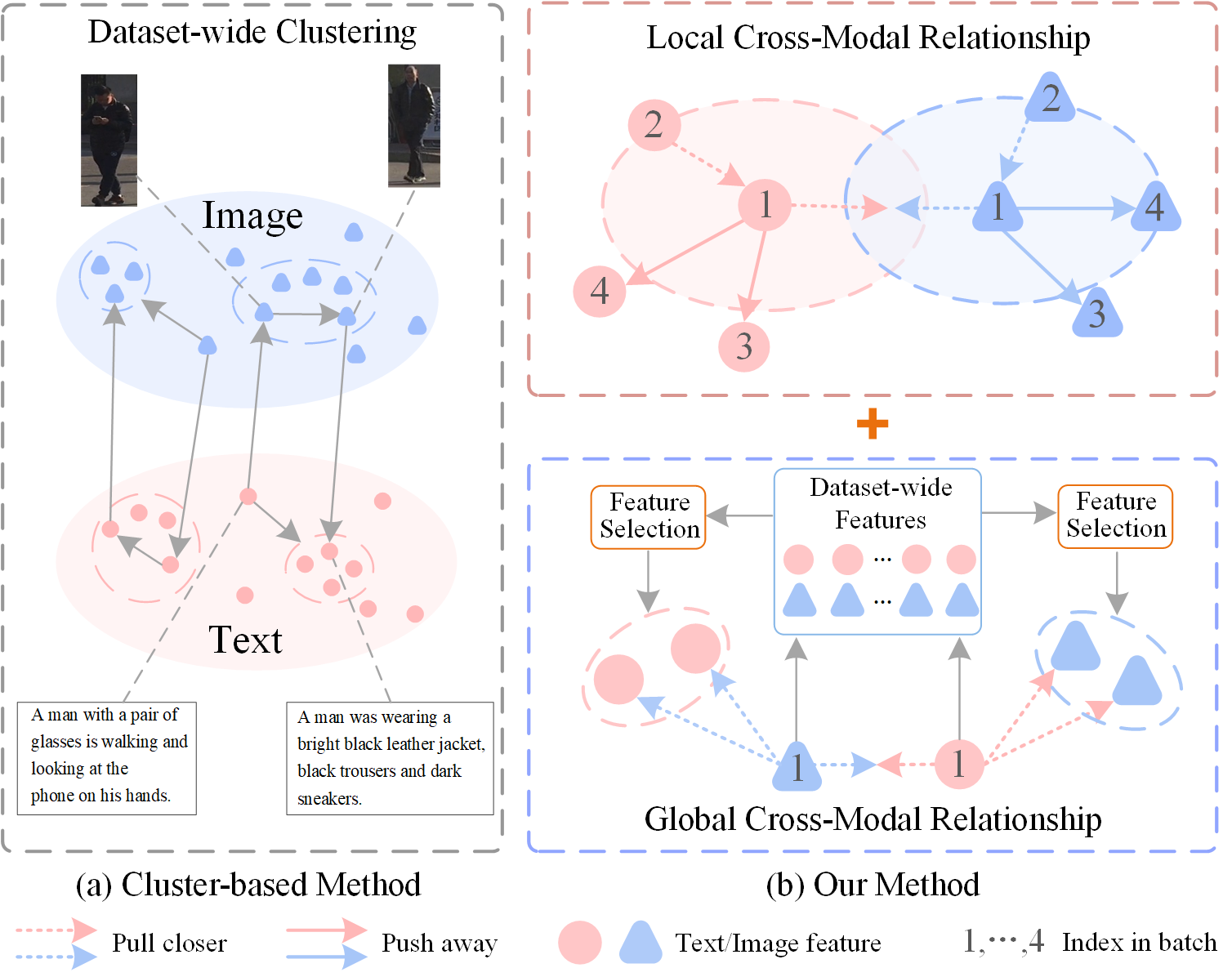}\vspace{-4.0mm}
	\caption{Comparison between the proposed method and existing approaches.
		While existing methods primarily rely on dataset-wide clustering to establish cross-modal identity associations, our method introduces a dual-granularity mechanism that integrates local and global collaboration, leading to more accurate and robust associations.}\vspace{-6mm}\label{Fig1} 
\end{figure}  

To address this issue, GTR \cite{10} and GAAP  \cite{9} leverage large image-text models to automatically generate textual descriptions for person images, thereby avoiding manual labeling of image-text pairs. However, the generated descriptions often contain noise, which weakens the model’s generalization performance and hinders further improvements in cross-modal matching. Weakly supervised text-to-person image matching assumes that each person image is paired with a textual description, while identity relationships between different images remain unknown. This setting not only eliminates the need for manual labeling of identity relationships but also reduces the noise introduced by unsupervised text generation. Nevertheless, the lack of complete cross-modal identity supervision significantly limits the model’s performance. To this end, methods such as CMMT \cite{7} and CPCL \cite{8} apply cross-modal clustering \cite{44} (as shown in Figure \ref{Fig1}(a)) to generate pseudo associations and train models in a supervised manner. Unfortunately, these methods struggle with hard samples, as clustering fails to effectively separate visually or semantically similar instances.

To address the aforementioned issues, this paper proposes a dual-granularity identity relationship modeling mechanism that operates at both local and global levels, as illustrated in Figure \ref{Fig1}(b). This mechanism enhances the model’s ability to distinguish hard samples by emphasizing those with uncertain identity relationships during training, thereby improving overall robustness. Specifically, the model first establishes identity correspondences among intra-modality samples and integrates them with cross-modal associations within each training batch. This facilitates the explicit construction of bidirectional cross-modal relationships— from text to image and image to text. Such modeling reinforces identity-level constraints across modalities, enhances the representational capacity of the base model, and provides a strong foundation for subsequent global relationship reasoning.

At the global relationship level, this paper addresses the heterogeneity of cross-modal data by using the visual modality as a relational anchor. Cross-modal identity associations are constructed by dynamically linking image samples with the entire training set. Leveraging the visual modality’s strength in fine-grained representation, image samples are treated as the “feature centers” of cross-modal relationships, helping to mitigate the ambiguity introduced by the semantic abstractness of textual descriptions. This enables the model to establish more reliable global identity associations. To further improve the model’s sensitivity to weak associations, a dynamic confidence adjustment mechanism is introduced, which nonlinearly adjusts confidence scores based on real-time feedback from inter-modal similarity responses. As a result, samples with weaker associations receive increased attention during feature extraction, encouraging the model to refine decision boundaries in ambiguous regions.  In addition, to compensate for the absence of labeled hard samples, we propose an information-asymmetric sample pair construction and consistency learning strategy. An information perturbation mechanism is applied to generate cross-modal sample pairs with semantic discrepancies, and a cross-modal semantic consistency regularization is employed to guide the model in learning robust representations under asymmetric conditions. This strategy not only addresses the challenge of mining hard cross-modal samples but also significantly enhances the model’s robustness in complex matching scenarios. Overall, the contributions of this paper are threefold: 
\vspace{-4.0mm}

\begin{itemize}
	\item We integrate identity relationships at both local and global levels by constructing local relationships within a single batch and establishing global cross-modal identity associations using the visual modality as an anchor. This mechanism significantly enhances the model's ability to distinguish hard samples and improves its overall robustness.
	\item We design a dynamic mechanism that nonlinearly adjusts relationship confidence based on real-time feedback from inter-modality similarity responses. This approach increases model's sensitivity to weakly associated samples and enables it to make finer distinctions near ambiguous decision boundaries.
	\item We propose an innovative strategy that addresses the absence of clearly labeled hard samples by generating cross-modal sample pairs with semantic discrepancies through an information perturbation strategy. Combined with cross-modality semantic consistency regularization, this method enables the model to learn robust feature representations under information-asymmetric conditions, significantly improving its robustness in complex matching scenarios.
\end{itemize}

\vspace{-3.0mm}
\section{Related Work}
\textbf{Supervised Methods}. In supervised text-to-person image matching, the identity correspondence between texts and person images is known. How to leverage these annotations to train models has been a research hotspot. In early studies, CNNs, such as VGG \cite{18} and ResNet \cite{19}, were commonly used to extract global representations from images, while LSTM \cite{17} or BERT \cite{16} was employed to extract global text representations, as seen in methods like MCCL \cite{21}, DCPL \cite{20}, and DCIE \cite{22}. The key to these methods lies in designing effective loss functions to enhance global cross-modal alignment. However, relying solely on global information makes it difficult to capture subtle visual cues of individuals, thereby limiting retrieval accuracy.

To address the challenges, explicit local alignment strategies have been proposed based on patch-to-word matching concepts \cite{23}. Methods like MIA \cite{25} and TIPCB \cite{24} adopt multi-stage or hierarchical frameworks to align horizontally segmented image regions with noun phrases, while PMA \cite{27} and ViTAA \cite{26} incorporate prior knowledge such as pose estimation and attribute segmentation to guide finer-grained partitioning. However, granularity mismatch between detailed visual inputs and coarse textual descriptions remains an issue, motivating strategies such as granularity unification and redundancy suppression \cite{14, 1, 29, 28, 11}. In contrast, implicit alignment methods like IRRA \cite{2} and SSAN \cite{15} utilize attention mechanisms or cross-modal interaction to learn local correspondences without explicit region division, offering greater flexibility. Recently, vision-language pre-trained models such as CLIP \cite{13} and ALBEF \cite{12} have gained significant attention, inspiring methods like LAIP \cite{6}, Rasa \cite{30}, WoRA \cite{31}, Cfine \cite{3}, PLOT \cite{4}, and LSPM \cite{5}, which leverage large-scale image-text pretraining to improve cross-modal representation and alignment via fine-tuning or adaptation. Despite their success, these models rely heavily on large labeled datasets, making it essential to develop methods that reduce labeling costs without significantly compromising performance.

\begin{figure*}[t!]
	\centering
	\includegraphics[width=0.91\linewidth]{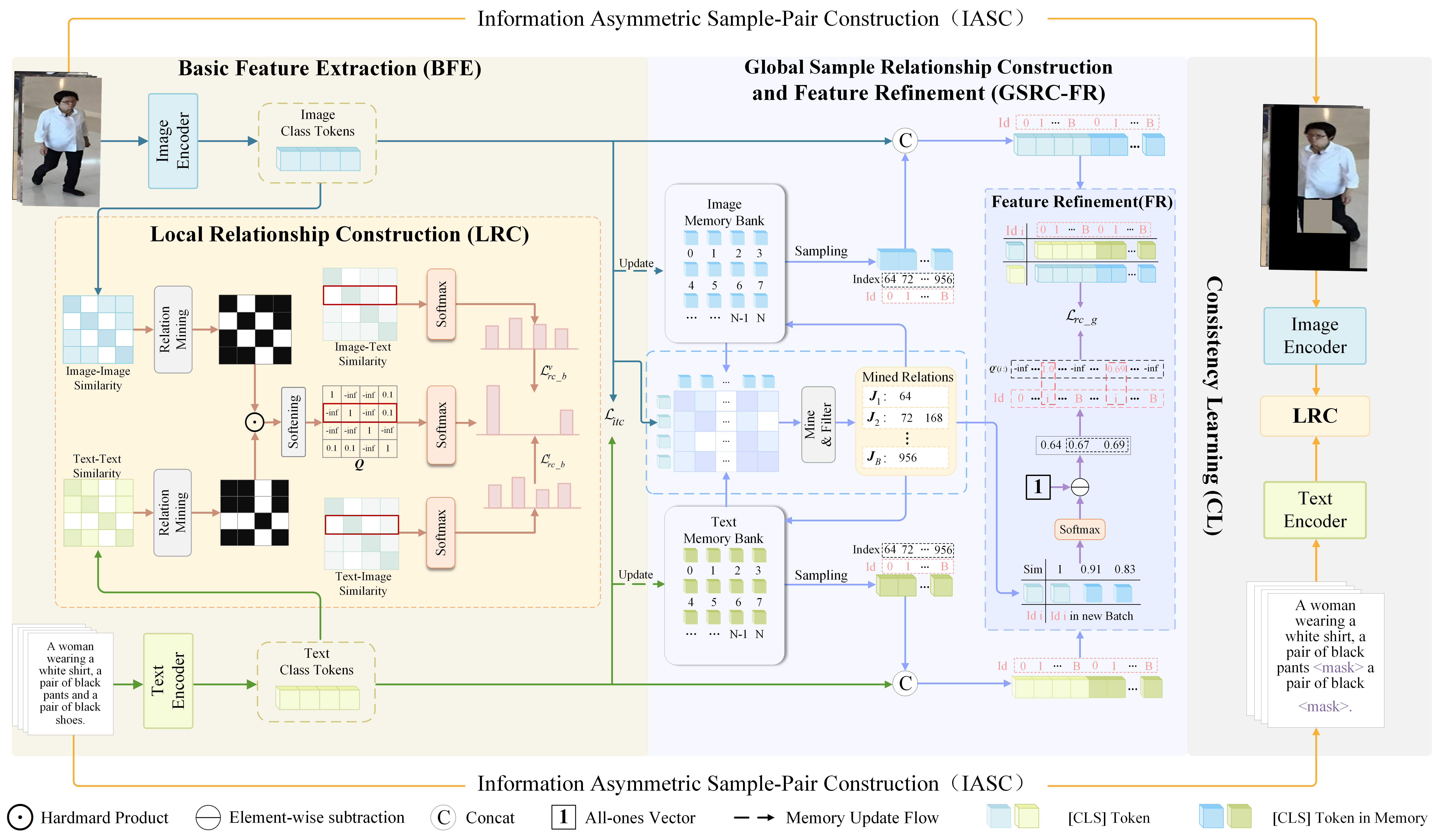}\vspace{-4.0mm}
	\caption{Overall architecture of the proposed method.   The input image-text pairs are first processed by the BFE module to extract basic representations and construct local cross-modal identity associations within each batch by integrating information from both modalities. The model then uses the visual modality as an anchor to explore broader cross-modal identity relationships from a global perspective. These global associations are leveraged to further optimize the model, improving its ability to distinguish between different identities. Additionally, information-asymmetric sample pairs are generated through specific transformations on the original inputs, enabling the model to better handle hard samples.}\label{Fig2}\vspace{-2mm}
\end{figure*}

\textbf{Unsupervised and Weakly Supervised Methods}. To reduce reliance on large-scale labeled samples, researchers have explored novel methods for unsupervised pedestrian text-image matching. Instead of assuming the availability of pre-existing textual descriptions of pedestrian appearance, these methods generate text descriptions directly from pedestrian images. The generated text-image pairs are then used for supervised training. Since no labeled training set is required, these approaches are classified as unsupervised text-to-person image matching methods. Among them, the quality of the generated text plays a crucial role in model performance. To address this issue, GAAP \cite{9} employs Attribute-guided Pseudo Caption Generation and incorporates an Attribute-guided Multi-strategy Cross-modal Alignment module to enhance cross-modal alignment. GTR \cite{10} generates fine-grained pseudo descriptions using a Vision-Language Model with instructional prompts and adopts a confidence-based training strategy during retrieval to mitigate the impact of noise in the generated text. Although these methods are effective, the noise in generated texts still limits the model’s generalization ability on real datasets.

In weakly supervised  setting, it is typically assumed that each person image has only one corresponding text description, while other correspondences remain unknown. This setup alleviates the challenge of cross-camera labeling for person images. However, relying solely on a single correspondence for training limits model performance. Therefore, in weakly supervised settings, constructing cross-modal relationships between samples and guiding model training accordingly is crucial. The earliest CMMT framework \cite{7} establishes cross-modal correspondences through clustering and improves accuracy via mutual training and relationship optimization. CPCL \cite{8} leverages CLIP’s strong cross-modal matching capabilities while using clustering to construct sample correspondences. However, these methods generally rely on clustering algorithms, such as DBSCAN \cite{44}, which makes it difficult to obtain precise cross-modal relationships, limiting model performance. Unlike these approaches, this paper introduces a mechanism that integrates intra-batch and global relationship construction to enhance the accuracy of cross-modal associations. By emphasizing samples with weak associations during training, it improves the model’s ability to distinguish difficult cases, ensuring robustness.

\section{Proposed Method}
\subsection{Problem Definition and Method Overview}
For the training set $\mathcal{D} = { (I_i, T_i) }_{i=1}^N$, where $T_i$ represents the $i$-th textual description, $I_i$ is the corresponding pedestrian image, and $N$ is the total number of image-text pairs. In the weakly supervised setting, a text-image pair $(I_i, T_j)$ shares the same pedestrian identity if $i = j$. When $i \neq j$, the identity correspondence between image $I_i$ and text $T_j$ is unknown. The key challenge in this setting lies in mining potential identity relationships among cross-modal samples based on known image-text associations. The overall architecture of the proposed method is illustrated in Figure~\ref{Fig2}. It consists of three key modules working collaboratively: Basic Feature Extraction (BFE), Global Sample Relationship Construction and Feature Refinement (GSRC-FR), and Information-Asymmetric Sample Pair Construction and Consistency Learning (IASC-CL). The BFE module integrates a text encoder and an image encoder, while reinforcing identity-level association constraints between the two modalities by explicitly constructing cross-modal relationships within a batch. This module establishes a solid foundation for GSRC-FR by jointly optimizing intra-modal feature representations and inter-modal identity alignment.

The GSRC-FR module automatically identifies sample pairs with clear identity matches based on global cross-modal correspondence and refines the features accordingly. The IASC-CL module addresses the absence of cross-modal hard samples by introducing an information perturbation strategy. Specifically, it applies operations such as feature masking and local region blocking to either the image or text modality to construct sample pairs with differing information across modalities. A contrastive learning paradigm is then employed to ensure the model maintains semantic consistency despite the introduced asymmetry. This mechanism effectively circumvents the difficulty of mining cross-modal hard samples and significantly enhances the model’s robustness in complex matching scenarios. Through the cascaded collaboration of these three modules, our method provides a discriminative and robust solution for cross-modal representation learning while fully leveraging the supervision from simple samples.

\subsection{Basic Feature Extraction}
\textbf{Feature Encoder}. For an input image $I_i$, we employ the CLIP \cite{13} pre-trained ViT-B/16 encoder to extract its global feature representation, following a mechanism similar to that of ViT \cite{49}. The resulting global image feature is denoted as $\bm{f}_i^v \in \mathbb{R}^{1 \times d}$. Likewise, for a given text description $T_i$, we use the CLIP pre-trained Transformer \cite{48} encoder to extract its global feature, denoted as $\bm{f}_i^t \in \mathbb{R}^{1 \times d}$. To enforce cross-modal consistency between $\bm{f}_i^v$ and $\bm{f}_i^t$, we adopt a contrastive learning objective \cite{45} that incorporates both a text-to-image loss $\mathcal{L}_{t2v}$ and an image-to-text loss $\mathcal{L}_{v2t}$. The text-to-image loss is formulated as:
\begin{equation}\small
	\mathcal{L}_{t2v} = - \frac{1}{B} \sum_{i=1}^B \log \frac{\exp (\text{sim}(\bm{f}_i^t, \bm{f}_i^v) / \tau)}{\sum_{k=1}^B \exp (\text{sim}(\bm{f}_i^t, \bm{f}_k^v) / \tau)}	
\end{equation}
where $\tau$ is a temperature coefficient that controls the sharpness of the probability distribution, $B$ denotes the batch size, and $\text{sim}(\bm{a}, \bm{b})$ denotes the cosine similarity between vectors $\bm{a}$ and $\bm{b}$. Similarly, the image-to-text loss $\mathcal{L}_{v2t}$ is defined in the same way. The total contrastive loss is given by:
\begin{equation}\small
	\mathcal{L}_{itc} = \mathcal{L}_{t2v} + \mathcal{L}_{v2t}
\end{equation}

\textbf{Local Relationship Construction}. In a weakly supervised setting, knowing only the correspondence between a single text and its paired image is insufficient for training a high-performance and robust model. To address this issue, an effective approach is to leverage the relationship between a given text and its corresponding image to mine potential identity correspondences across modalities. To this end, this paper proposes an intra-batch cross-modal relationship (i.e. local relationship) construction mechanism. Specifically, for a batch of $B$ paired samples, we compute the intra-modal self-similarity matrices $\bm{M}^v$ and $\bm{M}^t$. The element at position $(i,j)$ in $\bm{M}^v$ is defined as  
${\bm{M}^v}(i,j) = {\rm{sim}}( {\bm{f}_i^v}, \bm{f}_j^v)$.  
Similarly, the elements of $\bm{M}^t$ are given by  
${\bm{M}^t}(i,j) = {\rm{sim}}( {\bm{f}_i^t}, \bm{f}_j^t)$,  
where $i,j \in \{ 1, \cdots ,B\}$.

Through the similarity matrices $\bm{M}^v$ and $\bm{M}^t$, we can initially infer identity relationships among samples within the same modality. However, due to the influence of identity-irrelevant factors, $\bm{M}^v$ and $\bm{M}^t$ may not accurately reflect identity relationships. To address this, we refine $\bm{M}^v$ and $\bm{M}^t$ as follows:  
\begin{equation}\small
	{\tilde {\bm{M}}^k}(i,j) = \left\{ \begin{array}{ll}
		1 & \text{if } {\bm{M}^k}(i,j) > \text{th} \\
		0 & \text{if } {\bm{M}^k}(i,j) \le \text{th} \\
	\end{array} \right.
\end{equation}
where $k \in \{ v,t\}$ and $\text{th}$ is a threshold, set to 0.7 in this paper. Based on $\tilde {\bm{M}}^k$, we obtain the cross-modal identity correspondence matrix:  
\begin{equation}\small
	{\bm{M}^{v,t}} = {\tilde {\bm{M}}^v} \odot {\tilde {\bm{M}}^t}
\end{equation}
where $\odot$ denotes the Hadamard product. ${\bm{M}^{v,t}}$ integrates both intra-modal and cross-modal correlations among samples within a batch. When ${\bm{M}^{v,t}}(i,j) = 1$, it indicates a consistent identity correspondence between the $i$-th image and the $j$-th text; otherwise, no such correspondence exists. By establishing cross-modal correspondences, the proposed method not only fully utilizes the known correspondence between the $i$-th image and the $i$-th text but also incorporates intra-modal relationships, thereby enhancing the reliability of cross-modal relationship prediction.

To mitigate the significant negative impact of mismatched sample pairs on model training, we propose a relationship softening mechanism to reduce their adverse effects:
\begin{equation}\small
	\bm{Q} = \bm{I} + \lambda (\bm{M}^{v,t} - \bm{I})
\end{equation}
where $\bm{I}$ is the identity matrix and $0 < \lambda < 1$ serves as a balancing factor that controls the influence of mismatched correspondences during training. To further emphasize strong identity correspondences while suppressing weak ones, we refine $\bm{Q}$ as:
\begin{equation}\small
	\bm{Q}(i,j) = \left\{ \begin{array}{ll}
		-\infty & \text{if } \bm{Q}(i,j) = 0 \\
		\bm{Q}(i,j) & \text{otherwise} \\
	\end{array} \right.
\end{equation}
Based on the updated $\bm{Q}$, we define the target similarity probability $q_{i,j}$ as:
\begin{equation}\small
	q_{i,j} = \frac{\exp(\bm{Q}(i,j))}{\sum_{k=1}^B \exp(\bm{Q}(i,k))}
\end{equation}
The cross-modal similarity probability between the $i$-th image and the $j$-th text within a batch is given by:
\begin{equation}\small
	p_{i,j} = \frac{\exp\left(\text{sim}\left(\bm{f}_i^v, \bm{f}_j^t\right) / \tau\right)}{\sum_{k=1}^B \exp\left(\text{sim}\left(\bm{f}_i^v, \bm{f}_k^t\right) / \tau\right)}
\end{equation}

Guided by $q_{i,j}$, we optimize the representations $\bm{f}_i^v$ and $\bm{f}_i^t$ by aligning the similarity distribution $p_{i,j}$ with the target distribution $q_{i,j}$, thereby enhancing cross-modal matching capability. To achieve this, we introduce a similarity distribution matching (SDM) loss \cite{2, 20} to optimize the model parameters:
\begin{equation}\small
	\mathcal{L}_{rc\_b}^v = \frac{1}{B} \sum_{i=1}^B \sum_{j=1}^B p_{i,j} \log\left(\frac{p_{i,j}}{q_{i,j} + \varepsilon}\right)
\end{equation}
where $\varepsilon$ is a small constant added for numerical stability. Similarly, the within-batch relationship construction loss for text-to-image is defined as $\mathcal{L}_{rc\_b}^t$. The total local relationship construction loss is given by:
\begin{equation}\small
	\mathcal{L}_{rc\_b} = \mathcal{L}_{rc\_b}^v + \mathcal{L}_{rc\_b}^t
\end{equation}

\subsection{Global Relationship Construction and Feature Refinement}
Due to the limited number of samples within a single batch, there are relatively few cross-modal sample pairs with the same identity. To address this issue, we propose a GSRC-FR method. This approach constructs sample pairs with explicit cross-modal correspondences by leveraging the precise relationships between the current sample and same-modality samples from the entire training set. Since the sample relationships are built upon the entire dataset, we refer to this as a global sample construction method. To construct global sample relationships, we use the person image from the current batch as a visual anchor to measure its relevance with all samples in the dataset. For efficient implementation, we maintain Memory Banks for both modalities to store the representations of all training samples. Let ${\mathcal{B}^v} = \{ \hat{\bm{f}}_i^v \}_{i=1}^N$ and ${\mathcal{B}^t} = \{ \hat{\bm{f}}_i^t \}_{i=1}^N$ denote the Memory Banks for the image and text modalities, respectively. These Memory Banks are initialized using the representations $\bm{f}_i^v$ and $\bm{f}_i^t$ extracted by the encoders and are updated according to Eq.~(11). Taking the image Memory Bank as an example, the update rule is defined as:
\begin{equation}\small
	\hat{\bm{f}}_{i,n}^v = \alpha {\bm{f}}_i^v + (1 - \alpha) \hat{\bm{f}}_{i,n-1}^v
\end{equation}
where $\bm{f}_i^v$ is the newly extracted representation, $\hat{\bm{f}}_{i,n-1}^v$ is the stored representation from the previous epoch, $\hat{\bm{f}}_{i,n}^v$ is the updated representation after the $n$-th epoch, and $0 < \alpha < 1$ is a smoothing factor that mitigates abrupt changes in the stored features.

Assume $\bm{M}_m^v(i,j) = \text{sim}( \bm{f}_i^v, \hat{\bm{f}}_{j,n-1}^v )$, 
where $\bm{M}_m^v(i,j)$ denotes the similarity between the feature representation of the $i$-th pedestrian image in the current batch and that of the $j$-th image sample in the Memory Bank. 
Compared to textual descriptions, pedestrian images contain richer appearance information, which plays a more critical role in determining pedestrian identity. 
Therefore, we use the relationships among pedestrian images as a bridge to establish cross-modal associations in global relation modeling.  For each row of $\bm{M}_m^v$, we select the indices corresponding to the top $k$ highest similarity scores.  However, these top-$k$ samples may still include incorrectly matched instances. 
To mitigate this issue, we apply Eq.~(12) to filter out false matches from the top-$k$ candidates. The final set of selected indices is defined as:
\begin{equation}\small
	\bm{J}_i = \{ j \mid {\bm{M}}_m^v(i,j) > \text{th} \}, \quad i \in \{ 1, \ldots, B \}
\end{equation}
where $\text{th}$ is a threshold value, which is set to 0.7 in this paper. The pedestrian identities of the image samples indexed by $\bm{J}_i$ are regarded as being the same as that of the $i$-th image. 
We propagate this identity information to the text associated with the $i$-th image, thereby associating a single text with multiple pedestrian images. Meanwhile, we transfer the pedestrian identities carried by the texts corresponding to the pedestrian images in $\bm{J}_i$ back to the $i$-th image, thus linking a single pedestrian image with multiple pedestrian texts.

To effectively utilize cross-modal identity correspondences, we compute a similarity matrix $\bm{S}^{t,v}$ between the texts and pedestrian images within a batch, where $\bm{S}^{t,v}(i,j) = {\rm sim}(\bm{f}_i^t, \bm{f}_j^v)$ denotes the similarity between the $i$-th text and the $j$-th pedestrian image. In addition, we calculate the similarity between each text in the batch and the selected pedestrian images from the Memory Bank as $\bm{S}^{t,m}(i,j) = {\rm sim}(\bm{f}_i^t, \hat{\bm{f}}_{j,n-1}^v)$, where $j$ indexes samples in the sets $\{ \bm{J}_1, \dots, \bm{J}_B \}$. 
By concatenating $\bm{S}^{t,v}$ and $\bm{S}^{t,m}$ along the column dimension, we obtain $\bm{S}' = {\rm concat}(\bm{S}^{t,v}, \bm{S}^{t,m}) \in \mathbb{R}^{B \times B_1}$, where $B_1$ denotes the total number of pedestrian images in the current batch and the selected ones from the Memory Bank. 
Based on $\{ \bm{J}_1, \dots, \bm{J}_B \}$ and $\bm{S}'$, we construct a set $\bm{J}'$, which contains the column indices in $\bm{S}'$ corresponding to pedestrian images that share identity relationships with each text. 
Using $\bm{J}'$, we define the target relationship matrix $\bm{Q}' \in \mathbb{R}^{B \times B_1}$ as follows:


\begin{equation}\small
	\bm{Q}'(i,j) = \left\{ \begin{array}{cl}
		1 & \text{if } j \in \bm{J}'_i \\  
		-\infty & \text{otherwise}
	\end{array} \right.
\end{equation}

However, directly optimizing the model using $\bm{Q}'$ may cause it to focus predominantly on high-confidence sample pairs, while ignoring pairs with weaker but potentially informative correspondences. This could hinder the learning of more robust and generalizable representations. To address this issue, we propose an adaptive weight adjustment mechanism to regulate the influence of different sample pairs on the model. 
Specifically, for $j \in \bm{J}'_i$ and $j \ne i$, we update $\bm{Q}'(i,j)$ as follows:
\begin{equation}\small
	\bm{Q}'(i,j) = 1 - \frac{\exp({\rm sim}(\bm{f}_i^v, \bm{f}_j^v))}{\sum\limits_{k \in \bm{J}'_i} \exp({\rm sim}(\bm{f}_i^v, \bm{f}_k^v))}.
\end{equation}
To strengthen the associations between strongly correlated identity samples while suppressing the impact of weakly related pairs, we refine $\bm{Q}'$ using the weighting strategy defined above. 
With the updated $\bm{Q}'$, we compute the target similarity probability $q'_{i,j}$ as:
\begin{equation}\small
	q'_{i,j} = \frac{\exp(\bm{Q}'(i,j))}{\sum\limits_{k \in \bm{J}'_i} \exp(\bm{Q}'(i,k))}.
\end{equation}
Within the same batch, the predicted matching probability between the $i$-th text and the $j$-th image is given by:
\begin{equation}\small
	p'_{i,j} = \frac{\exp(\bm{S}'(i,j)/\tau)}{\sum\limits_{k=1}^{B_1} \exp(\bm{S}'(i,k)/\tau)}.
\end{equation}
Guided by $q'_{i,j}$, we refine the original within-batch representations $\bm{f}_i^v$ and $\bm{f}_i^t$ to incorporate richer identity-related information, thereby enhancing their cross-modal matching capability. 
To achieve this, we employ the SDM loss to optimize the model parameters:
\begin{equation}\small
	\mathcal{L}_{rc\_g}^{t} = \frac{1}{B} \sum\limits_{i=1}^B \sum\limits_{j=1}^{B_1} p'_{i,j} \log \left( \frac{p'_{i,j}}{q'_{i,j} + \epsilon} \right).
\end{equation}
Similarly, we define the global relation construction and representation refinement loss from image to text as $\mathcal{L}_{rc\_g}^{v}$. 
The final cross-modal global relation consistency loss is formulated as:
\begin{equation}\small
	\mathcal{L}_{rc\_g} = \mathcal{L}_{rc\_g}^{v} + \mathcal{L}_{rc\_g}^{t}.
\end{equation}
\subsection{Information Asymmetric Sample-Pair Construction and Consistency Learning}
Relying solely on the relationships within simple cross-modal sample pairs (i.e., between text and images) constructed in GSRC-FR is insufficient to address the complex and diverse cross-modal matching scenarios encountered in real-world applications. Directly mining cross-modal sample pairs from the training set for model training cannot ensure the correctness of the correspondences. To overcome this challenge, we propose the IASC-CL method to improve the model's cross-modal matching capability for hard samples. Specifically, given an image-text pair $(I_i, T_i)$, we define an image augmentation operator as:
\begin{equation}\small
	{\mathcal{A}}^v(I_i) = \text{Augment}(I_i)
\end{equation}
In Eq.~(19), the operator applies a series of transformations to the image, including horizontal flipping, border padding, random cropping, and random erasing. These augmentations simulate appearance variations caused by occlusion, viewpoint shifts, and other real-world factors. In parallel, we define a text augmentation operator as:
\begin{equation}\small
	{\mathcal{A}}^t(T_i) = \text{Mask}(T_i)
\end{equation}
Here, $\text{Mask}$ represents a masking operation that replaces parts of $T_i$ with \texttt{[MASK]}, simulating information loss due to incomplete descriptions or varying levels of granularity in the text.

The image-text pair constructed through Eqs.~(19) and (20), namely $({\mathcal{A}}^v(I_i), {\mathcal{A}}^t(T_i))$, exhibits weaker explicit semantic correlation compared to the original pair $(I_i, T_i)$, while still preserving underlying identity consistency. We therefore refer to these as information-asymmetric sample pairs, which are designed to simulate the complex and diverse cross-modal matching scenarios encountered in real-world applications. By exposing the model to such information-asymmetric pairs during training, it learns from more varied and challenging relationships, thereby enhancing its cross-modal matching capability in difficult cases. Similar to the within-batch cross-modal relationship construction loss, the consistency constraint for information-asymmetric sample pairs is formulated as:
\begin{equation}\small
	\mathcal{L}_{rc\_h} = \mathcal{L}_{rc\_h}^{\:v} + \mathcal{L}_{rc\_h}^{\:t}
\end{equation}
The overall training loss is defined as:
\begin{equation}\small
	\mathcal{L} = \mathcal{L}_{itc} + \mathcal{L}_{rc\_b} + \mathcal{L}_{rc\_g} + \mathcal{L}_{rc\_h}
\end{equation}
This formulation enables the model to not only handle straightforward matching cases, but also effectively tackle more complex scenarios involving information asymmetry, thereby improving its overall performance and discriminative capability.

\section{Experiments}
\subsection{Datasets and Evaluation Metrics}
\textbf{Datasets}. To evaluate the performance of our model, we use three widely used text-to-image person retrieval datasets: CUHK-PEDES \cite{32}, ICFG-PEDES \cite{15}, and RSTPReid \cite{14}.  

\textbf{CUHK-PEDES}. The dataset comprises 40,206 person images and 80,412 text descriptions corresponding to 13,003 identities. Each image is paired with at least two text descriptions, with an average length of at least 23 words. The dataset is divided into training, validation, and test sets. The training set contains 11,003 identities (34,054 images, 68,108 text descriptions), the validation set includes 1,000 identities (3,078 images, 6,158 text descriptions), and the test set consists of 1,000 identities (3,074 images, 6,156 text descriptions).  

\textbf{ICFG-PEDES}. The dataset consists of 54,522 images corresponding to 4,102 identities, with each image associated with a single text description. Compared to CUHK-PEDES, the text descriptions in ICFG-PEDES are more detailed, averaging 37 words in length. The dataset is split into a training set (34,674 image-text pairs, 3,102 identities) and a test set (19,848 image-text pairs, 1,000 identities).  

\textbf{RSTPReid}. The dataset is specifically designed for real-world applications and includes 20,505 images corresponding to 4,101 identities. Each identity has five images captured from different cameras, each paired with two text descriptions of at least 23 words. The training, validation, and test sets contain 3,701, 200, and 200 identities, respectively, corresponding to 18,505, 1,000, and 1,000 images, and 37,010, 2,000, and 2,000 text descriptions, respectively.  

\textbf{Evaluation Metrics}. To evaluate model performance, we use the Rank metric of the Cumulative Match Characteristic (CMC) \cite{42} for retrieval assessment. Additionally, we employ mean Average Precision (mAP) \cite{41} and mean Inverse Negative Penalty (mINP) \cite{40} as supplementary retrieval metrics. Higher values for these metrics indicate better retrieval performance.  

\subsection{Implementation Details}
The proposed method is implemented using the PyTorch framework, and all experiments are conducted on a single RTX 4090 GPU. The image encoder employs the CLIP-pretrained ViT-B/16 model, while the text encoder utilizes the CLIP-pretrained text Transformer. All input images are resized to $384\times128$ pixels, and the maximum text token sequence length is set to 77.  The batch size is set to 64, and training is conducted for 40 epochs using the Adam \cite{33} optimizer. The initial learning rate is set to $1\times10^{-5}$, following a cosine learning rate decay strategy. A 5-epoch warm-up \cite{34} period is applied at the beginning of training, during which the learning rate linearly increases from $1\times10^{-6}$ to $1\times10^{-5}$. The temperature parameter $\tau$ is fixed at 0.02.
\begin{table}[htbp]\footnotesize
	\centering
	\caption{Performance comparison of methods (\%) on the CUHK-PEDES dataset. The best performance in weakly supervised methods is highlighted in \textbf{bold}.}\vspace{-2.0mm}\label{Tab1}
	\resizebox{0.45\textwidth}{!}{ 
		\begin{tabular}{l|l|ccccc}
			\toprule
			Settings & Methods & R1 & R5 & R10 & mAP & mINP \\
			\midrule
			\multirow{9}{*}{Supervised} 
			& MANet~\cite{1} & 65.64 & 83.01 & 88.78 & -- & -- \\
			& LCR$^2$S~\cite{46} & 67.36 & 84.19 & 89.62 & 59.24 & -- \\
			& UniPT~\cite{37} & 68.50 & 84.67 & 90.38 & -- & -- \\
			& Cfine~\cite{3} & 69.57 & 85.93 & 91.15 & -- & -- \\
			& IA-PTIM~\cite{38} & 70.61 & 86.55 & 91.65 & -- & -- \\
			& IRRA~\cite{2} & 73.38 & 83.93 & 93.71 & 66.13 & 50.24 \\
			& LSPM~\cite{5} & 74.38 & 89.51 & 93.42 & 67.74 & 53.09 \\
			& PP-ReID\cite{47} & 74.89 & 89.90 & 94.17 & 67.12 & -- \\
			& PLOT~\cite{4} & 75.28 & 90.42 & 94.12 & -- & -- \\
			& APTM~\cite{36} & 76.53 & 90.04 & 94.15 & 66.91 & -- \\
			& LAIP~\cite{6} & 76.72 & 90.42 & 93.60 & 66.05 & -- \\
			& TriMatch ~\cite{39} & 76.84 & 89.90 & 94.17 & 67.12 & -- \\
			& PFM-EKFP~\cite{11} & 77.24 & 93.71 & 96.98 & 73.47 & -- \\
			& LUP-MLLM~\cite{35} & 78.13 & 91.19 & 94.50 & 68.75 & -- \\
			\midrule
			\multirow{2}{*}{Unsupervised} 
			& GTR~\cite{10} & 47.53 & 68.23 & 75.91 & 42.91 & -- \\
			& GAAP~\cite{9} & 47.64 & 67.79 & 76.08 & 41.28 & -- \\
			\midrule
			\multirow{3}{*}{Weakly Supervised} 
			& CMMT~\cite{7} & 57.10 & 78.14 & 85.23 & -- & -- \\
			& CPCL~\cite{8} & 70.03 & 87.28 & 91.78 & 63.19 & 47.54 \\
			& \textbf{Ours} & \textbf{73.06} & \textbf{89.21} & \textbf{93.44} & \textbf{64.88} & \textbf{48.15}\\
		    \bottomrule
		\end{tabular}
	}
\end{table}\vspace{-4.0mm}
\subsection{Comparison with State-of-the-Art Methods}
To validate the effectiveness of our method, we conduct comparisons with state-of-the-art (SOTA) methods in the supervised, weakly supervised, and unsupervised settings.

\textbf{Comparison with Supervised Methods}. As shown in Tables \ref{Tab1}-\ref{Tab3}, our weakly supervised method achieves competitive performance without relying on costly manual identity annotations. Although it still lags behind the most advanced supervised methods, it has already matched or even outperformed some recently published supervised approaches. This result demonstrates that our method effectively balances performance and annotation cost, making it more practical for real-world applications with limited annotation resources or budget constraints.
\begin{table}[h!]\footnotesize
	\centering
	\caption{Performance comparison of methods (\%) on the ICFG-PEDES dataset. The best performance in weakly supervised methods is highlighted in \textbf{bold}.}\label{Tab2}\vspace{-2.0mm}
	\resizebox{0.45\textwidth}{!}{ 
		\begin{tabular}{l|l|ccccc}
			\toprule
			\textbf{Labeled} & \textbf{Method} & \textbf{R1} & \textbf{R5} & \textbf{R10} & \textbf{mAP} & \textbf{mINP} \\
			\midrule
			\multirow{10}{*}{Supervised} 
			& LCR$^2$S\cite{46} & 57.93 & 76.08 & 82.40 & 38.21 & -- \\
			& MANet\cite{1} & 59.44 & 76.80 & 85.75 & -- & -- \\
		 	& IA-PTIM\cite{38} & 59.82 & 77.05 & 83.02 & -- & -- \\
			& UniPT\cite{37} & 60.09 & 76.19 & 82.46 & -- & -- \\
			& Cfine\cite{3} & 60.83 & 76.55 & 82.42 & -- & -- \\
			& IRRA\cite{2} & 63.46 & 80.25 & 85.82 & 38.06 & 7.93 \\
			& LAIP\cite{6} & 63.52 & 79.28 & 84.57 & 37.02 & -- \\
			& LSPM\cite{5} & 64.40 & 79.96 & 85.41 & 42.60 & 11.65 \\
			& PP-ReID\cite{47} & 65.12 & 81.57 & 86.97 & 42.93 & -- \\
			& PLOT\cite{4} & 65.76 & 81.39 & 86.73 & -- & -- \\
			& TriMatch\cite{39} & 67.71 & 85.37 & 88.02 & -- & -- \\
			& APTM\cite{36} & 68.51 & 82.99 & 87.56 & 41.22 & -- \\
			& PFM-EKFP\cite{11} & 69.29 & 89.10 & 94.06 & 47.15 & -- \\
			& LUP-MLLM\cite{35} & 69.37 & 83.55 & 88.18 & 42.42 & -- \\
			\cmidrule(r){1-7}
           \multirow{2}{*}{Unsupervised}   
           & GAAP\cite{9} & 27.12 & 44.91 & 53.56 & 11.43 & -- \\
           & GTR\cite{10} & 28.25 & 45.21 & 53.51 & 13.82 & -- \\
			\midrule
			\multirow{2}{*}{Weakly Supervised} & CPCL\cite{8} & 62.60 & 79.07 & 84.46 & 36.16 & \textbf{6.31} \\
			& Ours & \textbf{63.71} & \textbf{79.90} & \textbf{85.38} & \textbf{36.87} & 6.20 \\
			\bottomrule
		\end{tabular}\vspace{-4.0mm}
	}
\end{table}

\textbf{Comparison with Unsupervised Methods}. In the unsupervised setting, we omit manually annotated text and rely solely on image information, leveraging a large visual-language model (VLM) to generate text descriptions for training. During testing, we still use standard test sets with manually annotated text. GTR \cite{10} and GAAP \cite{9} are representative works in this field, aiming to optimize pseudo-text quality, but their performance remains limited. Our weakly supervised method outperforms the best unsupervised approaches by 25.42\%, 35.46\%, and 15.70\% in Rank-1 accuracy on the CUHK-PEDES, ICFG-PEDES, and RSTPReid datasets, respectively. This significant advantage highlights the limitations of current unsupervised methods and their gap from practical application goals. It further supports our view that weakly supervised methods offer a better trade-off between performance and cost-effectiveness at this stage. 

\begin{table}[htbp]\footnotesize
	\centering
	\caption{Performance Comparison of Methods (\%) on the RSTPReid dataset. The best performance in weakly supervised methods is highlighted in \textbf{bold}.}\vspace{-2.0mm}
	\label{Tab3}
	\resizebox{0.45\textwidth}{!}{ 
		\begin{tabular}{l|l|ccccc}
			\toprule
			\multirow{2}{*}{Labeled} & \multirow{2}{*}{Method} & \multicolumn{5}{c}{Performance Metrics} \\
			\cmidrule(r){3-7}
			& & R1 & R5 & R10 & mAP & mINP \\
			\midrule
			\multirow{8}{*}{Supervised} 
			& Cfine\cite{3} & 50.55 & 72.50 & 81.60 & -- & -- \\
			& UniPT\cite{37} & 51.85 & 74.85 & 82.85 & -- & -- \\
			& LCR$^2$S~\cite{46} & 54.95 & 76.65 & 84.70 & 40.92 & -- \\
			& IRRA\cite{2} & 60.20 & 81.30 & 88.20 & 47.17 & 25.28 \\
			& PLOT\cite{4} & 61.80 & 82.85 & 89.45 & -- & -- \\
			& PP-ReID\cite{47} & 61.87 & 83.63 & 89.70 & 47.82 & -- \\
			& LAIP\cite{6} & 62.00 & 83.15 & 88.50 & 45.27 & -- \\
			& IA-PTIM\cite{38} & 65.50 & 86.50 & 91.00 & -- & -- \\
			& TriMatch\cite{39} & 67.40 & 85.85 & 90.95 & -- & -- \\
			& APTM\cite{36} & 67.50 & 85.70 & 91.45 & 52.56 & -- \\
			& LUP-MLLM\cite{35} & 69.95 & 87.35 & 92.30 & 54.17 & -- \\
			\midrule
		    \multirow{2}{*}{Unsupervised} 
		    & GAAP\cite{9} & 44.45 & 65.15 & 75.30 & 31.21 & -- \\
		    & GTR\cite{10} & 45.60 & 70.35 & 79.95 & 33.30 & -- \\
			\midrule
			\multirow{2}{*}{Weakly Supervised} 
			& CPCL\cite{8} & 58.35 & 81.05 & 87.65 & 45.81 & 23.87 \\
			& Ours & \textbf{61.30} & \textbf{82.00} & \textbf{88.10} & \textbf{47.20} & \textbf{25.47}\\
			\bottomrule
		\end{tabular}
	}
\end{table}\vspace{-4.0mm}

\textbf{Comparison with Weakly Supervised Methods}. CMMT \cite{7} is a pioneering work in text-based person re-identification under weakly supervised conditions and reports results only on the CUHK-PEDES dataset. Therefore, we first validate our method on this benchmark. As shown in Table \ref{Tab1}, our method significantly outperforms existing weakly supervised methods, including CMMT \cite{7} and CPCL \cite{8}, achieving a Rank-1 accuracy of 73.06\% and an mAP of 64.88\%. Compared to CPCL, our method improves Rank-1 accuracy and mAP by 3.03\% and 1.69\%, respectively. 

Next, we evaluate our method on the ICFG-PEDES dataset, which contains more fine-grained text descriptions and imposes higher demands on text understanding and cross-modal matching. As shown in Table \ref{Tab2}, our method maintains strong performance, achieving a Rank-1 accuracy of 63.71\% and an mAP of 36.87\%, demonstrating its capability to handle complex and challenging text descriptions. Finally, on the diverse and realistic RSTPReid dataset, our method achieves a Rank-1 accuracy of 61.30\% and an mAP of 47.20\%. Compared to CPCL, it improves Rank-1 accuracy and mAP by 2.95 and 1.39 percentage points, respectively.

\subsection{Ablation Study}
In our method, we adopt the CLIP-ViT-B/16 model \cite{13} and fine-tune it using the loss function defined in Eq. (2) as the baseline for our approach. Our method primarily consists of three modules: BFE, GSRC-FR, and IASC-CL, with local relationship construction (LRC) being the core component of BFE. To validate the effectiveness of LRC, GSRC-FR, and IASC-CL, we conduct a series of ablation studies, and the results are presented in Table \ref{Tab4}. The experimental results show that when LRC is added to the baseline model, the Rank-1 accuracy on the CUHK-PEDES, ICFG-PEDES, and RSTPReid datasets improves by 4.35\%, 4.21\%, and 1.70\%, respectively. Similarly, when GSRC-FR is added to the baseline model, the Rank-1 accuracy increases by 2.87\%, 4.27\%, and 1.30\%, respectively. These results strongly demonstrate the effectiveness of LRC and GSRC-FR.

\begin{table}[htbp]\footnotesize
	\centering
	\caption{Ablation study on CUHK-PEDES, ICFG-PEDES, and RSTPReid datasets. `B' denotes the baseline method. Reported performance is Rank-1 accuracy.}\vspace{-2.0mm}
	\resizebox{0.45\textwidth}{!}{
	\begin{tabular}{l|ccc}
		\toprule
		Settings & CUHK-PEDES & ICFG-PEDES & RSTPReid \\
		\midrule
		B & 64.69 & 54.24 & 54.60 \\
		B+LRC & 69.04 & 58.45 & 56.30 \\
		B+GSRC-FR & 67.56 & 58.51 & 55.90 \\
		B+LRC+GSRC-FR & 70.20 & 60.30 & 57.25 \\
		B+LRC+GSRC-FR+IASC-CL & 73.06 & 63.71 & 61.30 \\
		\bottomrule
	\end{tabular}
	}\label{Tab4}
\end{table}\vspace{-4.0mm}

Furthermore, when GSRC-FR is added to the B+LRC model, the resulting model, B+LRC+GSRC-FR, achieves Rank-1 accuracy improvements of 5.51\%, 6.06\%, and 2.65\% on the three datasets compared to the baseline model, and improvements of 1.16\%, 1.85\%, and 0.95\% compared to the B+LRC model. This indicates that GSRC-FR effectively complements the limitations of LRC and further enhances the model's performance. To verify the effectiveness of IASC-CL, we add it to the B+LRC+GSRC-FR model, resulting in the final model, B+LRC+GSRC-FR+IASC-CL. As shown in Table \ref{Tab4}, compared to the B+LRC+GSRC-FR model, B+LRC+GSRC-FR+IASC-CL achieves Rank-1 accuracy improvements of 2.86\%, 3.41\%, and 4.05\% on the three datasets, respectively. Compared to the baseline model, the final model B+LRC+GSRC-FR+IASC-CL achieves Rank-1 accuracy improvements of 8.37\%, 9.47\%, and 6.70\% on the CUHK-PEDES, ICFG-PEDES, and RSTPReid datasets, respectively. These results fully demonstrate the effectiveness of each module and the advantages of their combined use.
\begin{table}[htbp]\footnotesize
	\centering
	\caption{Ablation study about BFE on CUHK-PEDES, ICFG-PEDES and RSTPReid datasets. `B' denotes the baseline method. Reported performance is Rank-1 accuracy.}\vspace{-2.0mm}
	\label{Tab5}	
	\resizebox{0.4\textwidth}{!}{
	\begin{tabular}{l|ccc}
		\toprule
		Settings & CUHK-PEDES & ICFG-PEDES & RSTPReid \\
		\midrule
		B & 64.69 & 54.24 & 54.60 \\
		B+LRCI & 67.56 & 51.32 & 53.51 \\
		B+LRCT & 67.61 & 56.47 & 55.45 \\
		B+LRC & 69.04 & 58.45 & 56.30 \\
		\bottomrule
	\end{tabular}}
\end{table}\vspace{-4.0mm}

\subsection{Further Discussion}
\textbf{Further Analysis of BFE}. 
In LRC, we integrate information from both image and text modalities based on Eq. (4) to determine cross-modal identity relationships within a batch. To verify the effectiveness of this integration approach, we conduct comparative experiments by constructing models that use only image information (LRCI) or only text information (LRCT) and evaluate their performance. As shown in Table \ref{Tab5}, when relying solely on single-modal relationships to establish cross-modal identity correspondences within a batch, although the model outperforms the baseline, its performance decreases to varying extents across different datasets compared to the B+LRC method, which utilizes both modalities. This is mainly because relying solely on single-modal relationships for cross-modal identity matching often results in inaccuracies and incompleteness, thereby limiting further performance improvements.\\

\begin{figure}[t!]\vspace{-2.0mm}
	\centering
	\includegraphics[width=0.75\linewidth]{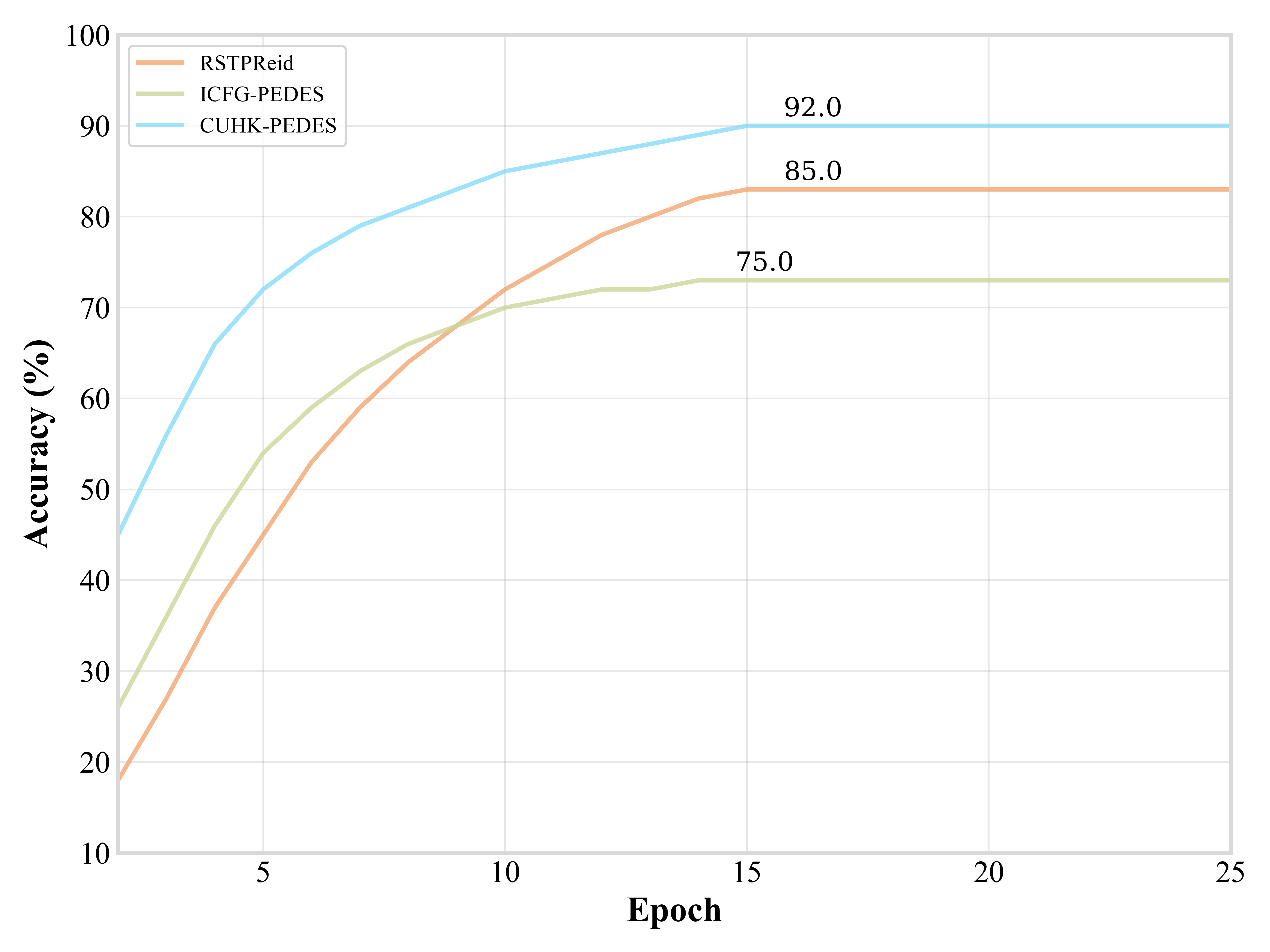}\vspace{-4.0mm}
	\caption{Accuracy of cross-modal relation constructed by GSCS-FR.}\label{Fig3}\vspace{-4mm}
\end{figure}

\begin{table}[htbp]\vspace{-2.0mm}\footnotesize
	\centering
	\caption{Ablation study about GSRC-FR on CUHK-PEDES, ICFG-PEDES and RSTPReid datasets. `B' denotes the baseline method. Reported performance is Rank-1 accuracy.}\vspace{-2.0mm}
	\label{Tab6}	
	\resizebox{0.4\textwidth}{!}{
		\begin{tabular}{l|ccc}
			\toprule
			Settings & CUHK-PEDES & ICFG-PEDES & RSTPReid \\
			\midrule
			BFE & 69.04 & 58.45 & 56.30 \\
			BFE+GSRC-T & 67.82 & 57.42 & 56.80 \\
			BFE+GSRC-IT & 68.73 & 58.22 & 56.90 \\
			BFE+GSRC-FR & 70.20 & 60.30 & 57.25 \\
			\bottomrule
	\end{tabular}}
\end{table}
\vspace{-2.0mm}

\textbf{Further Analysis of GSRC-FR}. Global relation construction serves as the core component of the GSRC-FR module. To evaluate its effectiveness, we compare its performance under three settings: a model using only BFE, a model constructing global identity relations using only text information (BFE+GSRC-T), and a model utilizing both image and text information (BFE+GSRC-IT). As shown in Table \ref{Tab6}, on the CUHK-PEDES and ICFG-PEDES datasets, the BFE+GSRC-T model performs worse than BFE, mainly because text information is less informative than pedestrian images, making it ineffective in constructing identity relations and thus hindering performance improvement. Compared to our proposed method, the BFE+GSRC-IT model fails to improve performance and instead suffers a performance drop. This is mainly because, while combining both image and text modalities helps identify more accurate identity relations, it also excludes training samples with uncertain relations, thereby reducing the model's generalization ability.

To further assess the effectiveness of GSRC-FR, we evaluate the accuracy of identity relations constructed by this module, using the ratio of correctly identified identities to the total number of mined identities as the metric. Figure \ref{Fig3} presents the experimental results. As the number of epoch increases, the accuracy of cross-modal identity relations predicted by BFE+GSRC-FR on the CUHK-PEDES dataset reaches 92.0\%. Even on the ICFG-PEDES dataset, where performance is relatively lower, the accuracy still reaches 75.0\%, which further demonstrates the effectiveness and reliability of the GSRC-FR module.
\begin{figure}[t!]
	\centering
	\includegraphics[width=1.0\linewidth]{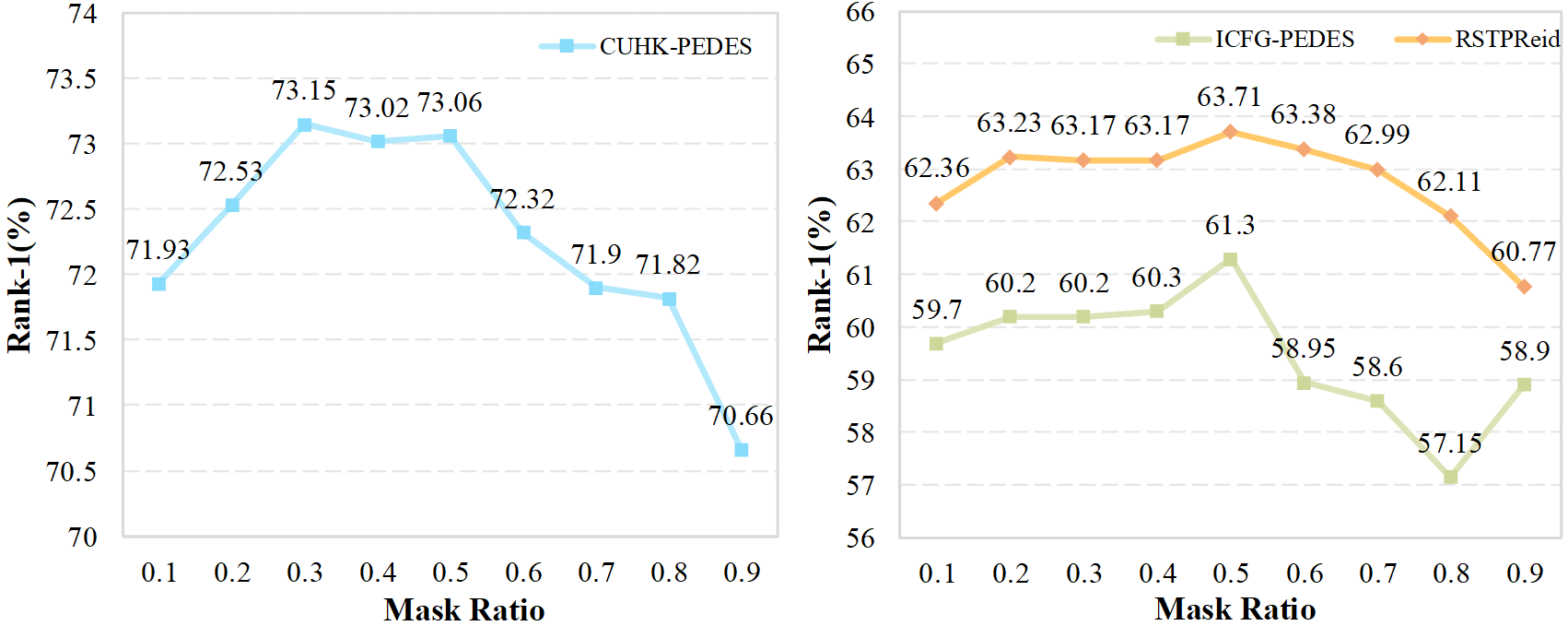}\vspace{-2.0mm}
	\caption{Analysis of the impact of masking ratio on model performance. (a) Effect of different masking ratios on model performance on the CUHK-PEDES dataset. (b) Effect of different masking ratios on model performance on the ICFG-PEDES and RSTPReid datasets.}\label{Fig4}\vspace{-2mm}
\end{figure}

\textbf{Further Analysis of IASC-CL}. In IASC-CL, applying different masking ratios to text affects model performance differently. To determine the optimal masking ratio, we apply different masking ratios to the text. As shown in Figure \ref{Fig4}, on the CUHK-PEDES dataset, model performance reaches its optimum when the masking ratio is between 0.3 and 0.5, whereas on the ICFG-PEDES and RSTPReid datasets, model performance is optimal when the masking ratio is between 0.4 and 0.5. Based on these results, we set the final masking ratio to 0.5.

\section{Conclusion}
This paper proposes a local-and-global dual-granularity identity association mechanism that effectively enhances text-to-person image matching under weakly supervised settings. By explicitly establishing cross-modal associations within each batch, identity constraints across modalities are reinforced, supporting global relationship reasoning. The dynamic association strategy, guided by visual information, alleviates ambiguity caused by abstract textual semantics, improving robustness. Additionally, the cross-modal confidence-based adaptive adjustment mechanism boosts sensitivity to weak associations, while the information-asymmetric sample pair construction effectively avoids the challenge of hard sample mining. Experimental results confirm the proposed method’s superiority and robustness on challenging benchmarks, promoting progress in intelligent security and smart city applications. Future work will explore multi-modal dynamic weight adaptation and cross-scenario generalization to further enhance practical deployment.

\bibliographystyle{ACM-Reference-Format}
\bibliography{acmart}


\begin{thebibliography}{49}


\ifx \showCODEN    \undefined \def \showCODEN     #1{\unskip}     \fi
\ifx \showISBNx    \undefined \def \showISBNx     #1{\unskip}     \fi
\ifx \showISBNxiii \undefined \def \showISBNxiii  #1{\unskip}     \fi
\ifx \showISSN     \undefined \def \showISSN      #1{\unskip}     \fi
\ifx \showLCCN     \undefined \def \showLCCN      #1{\unskip}     \fi
\ifx \shownote     \undefined \def \shownote      #1{#1}          \fi
\ifx \showarticletitle \undefined \def \showarticletitle #1{#1}   \fi
\ifx \showURL      \undefined \def \showURL       {\relax}        \fi
\providecommand\bibfield[2]{#2}
\providecommand\bibinfo[2]{#2}
\providecommand\natexlab[1]{#1}
\providecommand\showeprint[2][]{arXiv:#2}

\bibitem[Bai et~al\mbox{.}(2023a)]%
        {30}
\bibfield{author}{\bibinfo{person}{Yang Bai}, \bibinfo{person}{Min Cao},
  \bibinfo{person}{Daming Gao}, \bibinfo{person}{Ziqiang Cao},
  \bibinfo{person}{Chen Chen}, \bibinfo{person}{Zhenfeng Fan},
  \bibinfo{person}{Liqiang Nie}, {and} \bibinfo{person}{Min Zhang}.}
  \bibinfo{year}{2023}\natexlab{a}.
\newblock \showarticletitle{RaSa: Relation and Sensitivity Aware Representation
  Learning for Text-based Person Search}. In
  \bibinfo{booktitle}{\emph{Proceedings of the Thirty-Second International
  Joint Conference on Artificial Intelligence}}. \bibinfo{pages}{555--563}.
\newblock


\bibitem[Bai et~al\mbox{.}(2023b)]%
        {10}
\bibfield{author}{\bibinfo{person}{Yang Bai}, \bibinfo{person}{Jingyao Wang},
  \bibinfo{person}{Min Cao}, \bibinfo{person}{Chen Chen},
  \bibinfo{person}{Ziqiang Cao}, \bibinfo{person}{Liqiang Nie}, {and}
  \bibinfo{person}{Min Zhang}.} \bibinfo{year}{2023}\natexlab{b}.
\newblock \showarticletitle{Text-based person search without parallel
  image-text data}. In \bibinfo{booktitle}{\emph{Proceedings of the 31st ACM
  International Conference on Multimedia}}. \bibinfo{pages}{757--767}.
\newblock


\bibitem[Chen et~al\mbox{.}(2018)]%
        {23}
\bibfield{author}{\bibinfo{person}{Tianlang Chen}, \bibinfo{person}{Chenliang
  Xu}, {and} \bibinfo{person}{Jiebo Luo}.} \bibinfo{year}{2018}\natexlab{}.
\newblock \showarticletitle{Improving text-based person search by spatial
  matching and adaptive threshold}. In \bibinfo{booktitle}{\emph{2018 IEEE
  Winter Conference on Applications of Computer Vision (WACV)}}. IEEE,
  \bibinfo{pages}{1879--1887}.
\newblock


\bibitem[Chen et~al\mbox{.}(2022)]%
        {24}
\bibfield{author}{\bibinfo{person}{Yuhao Chen}, \bibinfo{person}{Guoqing
  Zhang}, \bibinfo{person}{Yujiang Lu}, \bibinfo{person}{Zhenxing Wang}, {and}
  \bibinfo{person}{Yuhui Zheng}.} \bibinfo{year}{2022}\natexlab{}.
\newblock \showarticletitle{TIPCB: A simple but effective part-based
  convolutional baseline for text-based person search}.
\newblock \bibinfo{journal}{\emph{Neurocomputing}}  \bibinfo{volume}{494}
  (\bibinfo{year}{2022}), \bibinfo{pages}{171--181}.
\newblock


\bibitem[Devlin et~al\mbox{.}(2019)]%
        {16}
\bibfield{author}{\bibinfo{person}{Jacob Devlin}, \bibinfo{person}{Ming-Wei
  Chang}, \bibinfo{person}{Kenton Lee}, {and} \bibinfo{person}{Kristina
  Toutanova}.} \bibinfo{year}{2019}\natexlab{}.
\newblock \showarticletitle{Bert: Pre-training of deep bidirectional
  transformers for language understanding}. In
  \bibinfo{booktitle}{\emph{Proceedings of the 2019 Conference of the North
  American Chapter of the Association for Computational Linguistics: Human
  Language Technologies, volume 1 (long and short papers)}}.
  \bibinfo{pages}{4171--4186}.
\newblock


\bibitem[Ding et~al\mbox{.}(2021)]%
        {15}
\bibfield{author}{\bibinfo{person}{Zefeng Ding}, \bibinfo{person}{Changxing
  Ding}, \bibinfo{person}{Zhiyin Shao}, {and} \bibinfo{person}{Dacheng Tao}.}
  \bibinfo{year}{2021}\natexlab{}.
\newblock \showarticletitle{Semantically self-aligned network for text-to-image
  part-aware person re-identification}.
\newblock \bibinfo{journal}{\emph{arXiv preprint arXiv:2107.12666}}
  (\bibinfo{year}{2021}).
\newblock


\bibitem[Dosovitskiy et~al\mbox{.}(2021)]%
        {49}
\bibfield{author}{\bibinfo{person}{Alexey Dosovitskiy}, \bibinfo{person}{Lucas
  Beyer}, \bibinfo{person}{Alexander Kolesnikov}, \bibinfo{person}{Dirk
  Weissenborn}, \bibinfo{person}{Xiaohua Zhai}, \bibinfo{person}{Thomas
  Unterthiner}, \bibinfo{person}{Mostafa Dehghani}, \bibinfo{person}{Matthias
  Minderer}, \bibinfo{person}{Georg Heigold}, \bibinfo{person}{Sylvain Gelly},
  \bibinfo{person}{Jakob Uszkoreit}, {and} \bibinfo{person}{Neil Houlsby}.}
  \bibinfo{year}{2021}\natexlab{}.
\newblock \showarticletitle{An Image is Worth 16x16 Words: Transformers for
  Image Recognition at Scale}.
\newblock \bibinfo{journal}{\emph{ICLR}} (\bibinfo{year}{2021}).
\newblock


\bibitem[Ester et~al\mbox{.}(1996)]%
        {44}
\bibfield{author}{\bibinfo{person}{Martin Ester}, \bibinfo{person}{Hans-Peter
  Kriegel}, \bibinfo{person}{J{\"o}rg Sander}, {and} \bibinfo{person}{Xiaowei
  Xu}.} \bibinfo{year}{1996}\natexlab{}.
\newblock \showarticletitle{A density-based algorithm for discovering clusters
  in large spatial databases with noise}. In
  \bibinfo{booktitle}{\emph{Proceedings of the Second International Conference
  on Knowledge Discovery and Data Mining}}. \bibinfo{pages}{226--231}.
\newblock


\bibitem[He et~al\mbox{.}(2016)]%
        {19}
\bibfield{author}{\bibinfo{person}{Kaiming He}, \bibinfo{person}{X. Zhang},
  \bibinfo{person}{Shaoqing Ren}, {and} \bibinfo{person}{Jian Su}.}
  \bibinfo{year}{2016}\natexlab{}.
\newblock \showarticletitle{Deep Residual Learning for Image Recognition}. In
  \bibinfo{booktitle}{\emph{2016 IEEE Conference on Computer Vision and Pattern
  Recognition (CVPR)}}. \bibinfo{pages}{770--778}.
\newblock


\bibitem[Hochreiter and Schmidhuber(1997)]%
        {17}
\bibfield{author}{\bibinfo{person}{Sepp Hochreiter} {and}
  \bibinfo{person}{J{\"u}rgen Schmidhuber}.} \bibinfo{year}{1997}\natexlab{}.
\newblock \showarticletitle{Long short-term memory}.
\newblock \bibinfo{journal}{\emph{Neural Computation}} \bibinfo{volume}{9},
  \bibinfo{number}{8} (\bibinfo{year}{1997}), \bibinfo{pages}{1735--1780}.
\newblock


\bibitem[Jiang and Ye(2023)]%
        {2}
\bibfield{author}{\bibinfo{person}{Ding Jiang} {and} \bibinfo{person}{Mang
  Ye}.} \bibinfo{year}{2023}\natexlab{}.
\newblock \showarticletitle{Cross-modal implicit relation reasoning and
  aligning for text-to-image person retrieval}. In
  \bibinfo{booktitle}{\emph{Proceedings of the IEEE/CVF Conference on Computer
  Vision and Pattern Recognition}}. \bibinfo{pages}{2787--2797}.
\newblock


\bibitem[Jing et~al\mbox{.}(2020)]%
        {27}
\bibfield{author}{\bibinfo{person}{Ya Jing}, \bibinfo{person}{Chenyang Si},
  \bibinfo{person}{Junbo Wang}, \bibinfo{person}{Wei Wang},
  \bibinfo{person}{Liang Wang}, {and} \bibinfo{person}{Tieniu Tan}.}
  \bibinfo{year}{2020}\natexlab{}.
\newblock \showarticletitle{Pose-guided multi-granularity attention network for
  text-based person search}. In \bibinfo{booktitle}{\emph{Proceedings of the
  AAAI Conference on Artificial Intelligence}}, Vol.~\bibinfo{volume}{34}.
  \bibinfo{pages}{11189--11196}.
\newblock


\bibitem[Kingma and Ba(2015)]%
        {33}
\bibfield{author}{\bibinfo{person}{Diederik~P. Kingma} {and}
  \bibinfo{person}{Jimmy Ba}.} \bibinfo{year}{2015}\natexlab{}.
\newblock \showarticletitle{Adam: {A} Method for Stochastic Optimization}. In
  \bibinfo{booktitle}{\emph{{ICLR}}}.
\newblock


\bibitem[Li et~al\mbox{.}(2025b)]%
        {38}
\bibfield{author}{\bibinfo{person}{Fan Li}, \bibinfo{person}{Hang Zhou},
  \bibinfo{person}{Huafeng Li}, \bibinfo{person}{Yafei Zhang}, {and}
  \bibinfo{person}{Zhengtao Yu}.} \bibinfo{year}{2025}\natexlab{b}.
\newblock \showarticletitle{Person text-image matching via text-feature
  interpretability embedding and external attack node implantation}.
\newblock \bibinfo{journal}{\emph{IEEE Transactions on Emerging Topics in
  Computational Intelligence}} \bibinfo{volume}{9}, \bibinfo{number}{2}
  (\bibinfo{year}{2025}), \bibinfo{pages}{1202--1215}.
\newblock


\bibitem[Li et~al\mbox{.}(2025a)]%
        {11}
\bibfield{author}{\bibinfo{person}{Huafeng Li}, \bibinfo{person}{Shedan Yang},
  \bibinfo{person}{Yafei Zhang}, \bibinfo{person}{Dapeng Tao}, {and}
  \bibinfo{person}{Zhengtao Yu}.} \bibinfo{year}{2025}\natexlab{a}.
\newblock \showarticletitle{Progressive Feature Mining and External
  Knowledge-Assisted Text-Pedestrian Image Retrieval}.
\newblock \bibinfo{journal}{\emph{IEEE Transactions on Multimedia}}
  \bibinfo{volume}{27} (\bibinfo{year}{2025}), \bibinfo{pages}{1973--1987}.
\newblock


\bibitem[Li et~al\mbox{.}(2024a)]%
        {5}
\bibfield{author}{\bibinfo{person}{Jiayi Li}, \bibinfo{person}{Min Jiang},
  \bibinfo{person}{Jun Kong}, \bibinfo{person}{Xuefeng Tao}, {and}
  \bibinfo{person}{Xi Luo}.} \bibinfo{year}{2024}\natexlab{a}.
\newblock \showarticletitle{Learning semantic polymorphic mapping for
  text-based person retrieval}.
\newblock \bibinfo{journal}{\emph{IEEE Transactions on Multimedia}}
  \bibinfo{volume}{26} (\bibinfo{year}{2024}), \bibinfo{pages}{10678--10691}.
\newblock


\bibitem[Li et~al\mbox{.}(2021)]%
        {12}
\bibfield{author}{\bibinfo{person}{Junnan Li}, \bibinfo{person}{Ramprasaath
  Selvaraju}, \bibinfo{person}{Akhilesh Gotmare}, \bibinfo{person}{Shafiq
  Joty}, \bibinfo{person}{Caiming Xiong}, {and} \bibinfo{person}{Steven
  Chu~Hong Hoi}.} \bibinfo{year}{2021}\natexlab{}.
\newblock \showarticletitle{Align before fuse: Vision and language
  representation learning with momentum distillation}. In
  \bibinfo{booktitle}{\emph{Proceedings of the 35th International Conference on
  Neural Information Processing Systems}}. \bibinfo{pages}{9694--9705}.
\newblock


\bibitem[Li et~al\mbox{.}(2017)]%
        {32}
\bibfield{author}{\bibinfo{person}{Shuang Li}, \bibinfo{person}{Tong Xiao},
  \bibinfo{person}{Hongsheng Li}, \bibinfo{person}{Bolei Zhou},
  \bibinfo{person}{Dayu Yue}, {and} \bibinfo{person}{Xiaogang Wang}.}
  \bibinfo{year}{2017}\natexlab{}.
\newblock \showarticletitle{Person Search with Natural Language Description}.
\newblock \bibinfo{journal}{\emph{2017 IEEE Conference on Computer Vision and
  Pattern Recognition (CVPR)}} (\bibinfo{year}{2017}),
  \bibinfo{pages}{5187--5196}.
\newblock


\bibitem[Li et~al\mbox{.}(2024b)]%
        {9}
\bibfield{author}{\bibinfo{person}{Zongyi Li}, \bibinfo{person}{Jianbo Li},
  \bibinfo{person}{Yuxuan Shi}, \bibinfo{person}{Hefei Ling},
  \bibinfo{person}{Jiazhong Chen}, \bibinfo{person}{Runsheng Wang}, {and}
  \bibinfo{person}{Shijuan Huang}.} \bibinfo{year}{2024}\natexlab{b}.
\newblock \showarticletitle{Cross-modal generation and alignment via
  attribute-guided prompt for unsupervised text-based person retrieval}. In
  \bibinfo{booktitle}{\emph{Proceedings of the Thirty-Third International Joint
  Conference on Artificial Intelligence}}. \bibinfo{pages}{1047--1055}.
\newblock


\bibitem[Luo et~al\mbox{.}(2019)]%
        {34}
\bibfield{author}{\bibinfo{person}{Hao Luo}, \bibinfo{person}{Youzhi Gu},
  \bibinfo{person}{Xingyu Liao}, \bibinfo{person}{Shenqi Lai}, {and}
  \bibinfo{person}{Wei Jiang}.} \bibinfo{year}{2019}\natexlab{}.
\newblock \showarticletitle{Bag of tricks and a strong baseline for deep person
  re-identification}. In \bibinfo{booktitle}{\emph{Proceedings of the IEEE/CVF
  Conference on Computer Vision and Pattern Recognition Workshops}}.
  \bibinfo{pages}{1487--1495}.
\newblock


\bibitem[Niu et~al\mbox{.}(2020)]%
        {25}
\bibfield{author}{\bibinfo{person}{Kai Niu}, \bibinfo{person}{Yan Huang},
  \bibinfo{person}{Wanli Ouyang}, {and} \bibinfo{person}{Liang Wang}.}
  \bibinfo{year}{2020}\natexlab{}.
\newblock \showarticletitle{Improving description-based person
  re-identification by multi-granularity image-text alignments}.
\newblock \bibinfo{journal}{\emph{IEEE Transactions on Image Processing}}
  \bibinfo{volume}{29} (\bibinfo{year}{2020}), \bibinfo{pages}{5542--5556}.
\newblock


\bibitem[Oord et~al\mbox{.}(2018)]%
        {45}
\bibfield{author}{\bibinfo{person}{Aaron van~den Oord}, \bibinfo{person}{Yazhe
  Li}, {and} \bibinfo{person}{Oriol Vinyals}.} \bibinfo{year}{2018}\natexlab{}.
\newblock \showarticletitle{Representation learning with contrastive predictive
  coding}.
\newblock \bibinfo{journal}{\emph{arXiv preprint arXiv:1807.03748}}
  (\bibinfo{year}{2018}).
\newblock


\bibitem[Park et~al\mbox{.}(2024)]%
        {4}
\bibfield{author}{\bibinfo{person}{Jicheol Park}, \bibinfo{person}{Dongwon
  Kim}, \bibinfo{person}{Boseung Jeong}, {and} \bibinfo{person}{Suha Kwak}.}
  \bibinfo{year}{2024}\natexlab{}.
\newblock \showarticletitle{PLOT: Text-based person search with part slot
  attention for corresponding part discovery}. In
  \bibinfo{booktitle}{\emph{Computer Vision – ECCV 2024: 18th European
  Conference}}. Springer, \bibinfo{pages}{474--490}.
\newblock


\bibitem[Radford et~al\mbox{.}(2021)]%
        {13}
\bibfield{author}{\bibinfo{person}{Alec Radford}, \bibinfo{person}{Jong~Wook
  Kim}, \bibinfo{person}{Chris Hallacy}, \bibinfo{person}{Aditya Ramesh},
  \bibinfo{person}{Gabriel Goh}, \bibinfo{person}{Sandhini Agarwal},
  \bibinfo{person}{Girish Sastry}, \bibinfo{person}{Amanda Askell},
  \bibinfo{person}{Pamela Mishkin}, \bibinfo{person}{Jack Clark},
  {et~al\mbox{.}}} \bibinfo{year}{2021}\natexlab{}.
\newblock \showarticletitle{Learning transferable visual models from natural
  language supervision}. In \bibinfo{booktitle}{\emph{Proceedings of the 38th
  International Conference on Machine Learning}}. PMLR,
  \bibinfo{pages}{8748--8763}.
\newblock


\bibitem[Shao et~al\mbox{.}(2023)]%
        {37}
\bibfield{author}{\bibinfo{person}{Zhiyin Shao}, \bibinfo{person}{Xinyu Zhang},
  \bibinfo{person}{Changxing Ding}, \bibinfo{person}{Jian Wang}, {and}
  \bibinfo{person}{Jingdong Wang}.} \bibinfo{year}{2023}\natexlab{}.
\newblock \showarticletitle{Unified pre-training with pseudo texts for
  text-to-image person re-identification}. In
  \bibinfo{booktitle}{\emph{Proceedings of the IEEE/CVF International
  Conference on Computer Vision}}. \bibinfo{pages}{11174--11184}.
\newblock


\bibitem[Shao et~al\mbox{.}(2022)]%
        {29}
\bibfield{author}{\bibinfo{person}{Zhiyin Shao}, \bibinfo{person}{Xinyu Zhang},
  \bibinfo{person}{Meng Fang}, \bibinfo{person}{Zhifeng Lin},
  \bibinfo{person}{Jian Wang}, {and} \bibinfo{person}{Changxing Ding}.}
  \bibinfo{year}{2022}\natexlab{}.
\newblock \showarticletitle{Learning granularity-unified representations for
  text-to-image person re-identification}. In
  \bibinfo{booktitle}{\emph{Proceedings of the 30th ACM International
  Conference on Multimedia}}. \bibinfo{pages}{5566--5574}.
\newblock


\bibitem[Simonyan and Zisserman(2015)]%
        {18}
\bibfield{author}{\bibinfo{person}{Karen Simonyan} {and}
  \bibinfo{person}{Andrew Zisserman}.} \bibinfo{year}{2015}\natexlab{}.
\newblock \showarticletitle{Very Deep Convolutional Networks for Large-Scale
  Image Recognition}. In \bibinfo{booktitle}{\emph{{ICLR}}}.
\newblock


\bibitem[Subramanyam et~al\mbox{.}(2023)]%
        {43}
\bibfield{author}{\bibinfo{person}{A~Venkata Subramanyam},
  \bibinfo{person}{Vibhu Dubey}, \bibinfo{person}{Niranjan Sundararajan}, {and}
  \bibinfo{person}{Brejesh Lall}.} \bibinfo{year}{2023}\natexlab{}.
\newblock \showarticletitle{Dense captioning for Text-Image ReID}. In
  \bibinfo{booktitle}{\emph{Proceedings of the Fourteenth Indian Conference on
  Computer Vision, Graphics and Image Processing}}. \bibinfo{pages}{1--8}.
\newblock


\bibitem[Sun et~al\mbox{.}(2024)]%
        {31}
\bibfield{author}{\bibinfo{person}{Jintao Sun}, \bibinfo{person}{Hao Fei},
  \bibinfo{person}{Zhedong Zheng}, {and} \bibinfo{person}{Gangyi Ding}.}
  \bibinfo{year}{2024}\natexlab{}.
\newblock \showarticletitle{From Data Deluge to Data Curation: A Filtering-WoRA
  Paradigm for Efficient Text-based Person Search}.
\newblock \bibinfo{journal}{\emph{arXiv preprint arXiv:2404.10292}}
  (\bibinfo{year}{2024}).
\newblock


\bibitem[Tan et~al\mbox{.}(2024)]%
        {35}
\bibfield{author}{\bibinfo{person}{Wentan Tan}, \bibinfo{person}{Changxing
  Ding}, \bibinfo{person}{Jiayu Jiang}, \bibinfo{person}{Fei Wang},
  \bibinfo{person}{Yibing Zhan}, {and} \bibinfo{person}{Dapeng Tao}.}
  \bibinfo{year}{2024}\natexlab{}.
\newblock \showarticletitle{Harnessing the power of mllms for transferable
  text-to-image person reid}. In \bibinfo{booktitle}{\emph{Proceedings of the
  IEEE/CVF Conference on Computer Vision and Pattern Recognition}}.
  \bibinfo{pages}{17127--17137}.
\newblock


\bibitem[Vaswani et~al\mbox{.}(2017)]%
        {48}
\bibfield{author}{\bibinfo{person}{Ashish Vaswani}, \bibinfo{person}{Noam
  Shazeer}, \bibinfo{person}{Niki Parmar}, \bibinfo{person}{Jakob Uszkoreit},
  \bibinfo{person}{Llion Jones}, \bibinfo{person}{Aidan~N Gomez},
  \bibinfo{person}{{\L}ukasz Kaiser}, {and} \bibinfo{person}{Illia
  Polosukhin}.} \bibinfo{year}{2017}\natexlab{}.
\newblock \showarticletitle{Attention is all you need}. In
  \bibinfo{booktitle}{\emph{Proceedings of the 31st International Conference on
  Neural Information Processing Systems}}. \bibinfo{pages}{6000--6010}.
\newblock


\bibitem[Wang et~al\mbox{.}(2021)]%
        {28}
\bibfield{author}{\bibinfo{person}{Chengji Wang}, \bibinfo{person}{Zhiming
  Luo}, \bibinfo{person}{Yaojin Lin}, {and} \bibinfo{person}{Shaozi Li}.}
  \bibinfo{year}{2021}\natexlab{}.
\newblock \showarticletitle{Text-based person search via multi-granularity
  embedding learning}. In \bibinfo{booktitle}{\emph{Proceedings of the
  Thirtieth International Joint Conference on Artificial Intelligence}}.
  \bibinfo{pages}{1068--1074}.
\newblock


\bibitem[Wang et~al\mbox{.}(2007)]%
        {42}
\bibfield{author}{\bibinfo{person}{Xiaogang Wang}, \bibinfo{person}{Gianfranco
  Doretto}, \bibinfo{person}{Thomas Sebastian}, \bibinfo{person}{Jens
  Rittscher}, {and} \bibinfo{person}{Peter Tu}.}
  \bibinfo{year}{2007}\natexlab{}.
\newblock \showarticletitle{Shape and appearance context modeling}. In
  \bibinfo{booktitle}{\emph{2007 IEEE 11th International Conference on Computer
  Vision}}. IEEE, \bibinfo{pages}{1--8}.
\newblock


\bibitem[Wang et~al\mbox{.}(2019)]%
        {21}
\bibfield{author}{\bibinfo{person}{Yuyu Wang}, \bibinfo{person}{Chunjuan Bo},
  \bibinfo{person}{Dong Wang}, \bibinfo{person}{Shuang Wang},
  \bibinfo{person}{Yunwei Qi}, {and} \bibinfo{person}{Huchuan Lu}.}
  \bibinfo{year}{2019}\natexlab{}.
\newblock \showarticletitle{Language person search with mutually connected
  classification loss}. In \bibinfo{booktitle}{\emph{ICASSP 2019-2019 IEEE
  International Conference on Acoustics, Speech and Signal Processing
  (ICASSP)}}. IEEE, \bibinfo{pages}{2057--2061}.
\newblock


\bibitem[Wang et~al\mbox{.}(2020)]%
        {26}
\bibfield{author}{\bibinfo{person}{Zhe Wang}, \bibinfo{person}{Zhiyuan Fang},
  \bibinfo{person}{Jun Wang}, {and} \bibinfo{person}{Yezhou Yang}.}
  \bibinfo{year}{2020}\natexlab{}.
\newblock \showarticletitle{Vitaa: Visual-textual attributes alignment in
  person search by natural language}. In \bibinfo{booktitle}{\emph{Computer
  vision--ECCV 2020: 16th European Conference}}. Springer,
  \bibinfo{pages}{402--420}.
\newblock


\bibitem[Wu et~al\mbox{.}(2024)]%
        {6}
\bibfield{author}{\bibinfo{person}{Yu Wu}, \bibinfo{person}{Haiguang Wang},
  \bibinfo{person}{Mengxia Wu}, \bibinfo{person}{Min Cao}, {and}
  \bibinfo{person}{Min Zhang}.} \bibinfo{year}{2024}\natexlab{}.
\newblock \showarticletitle{LAIP: learning local alignment from image-phrase
  modeling for text-based person search}. In \bibinfo{booktitle}{\emph{2024
  IEEE International Conference on Multimedia and Expo (ICME)}}. IEEE,
  \bibinfo{pages}{1--10}.
\newblock


\bibitem[Yan et~al\mbox{.}(2025)]%
        {39}
\bibfield{author}{\bibinfo{person}{Shuanglin Yan}, \bibinfo{person}{Neng Dong},
  \bibinfo{person}{Shuang Li}, {and} \bibinfo{person}{Huafeng Li}.}
  \bibinfo{year}{2025}\natexlab{}.
\newblock \showarticletitle{TriMatch: Triple Matching for Text-to-Image Person
  Re-Identification}.
\newblock \bibinfo{journal}{\emph{IEEE Signal Processing Letters}}
  \bibinfo{volume}{32} (\bibinfo{year}{2025}), \bibinfo{pages}{806--810}.
\newblock


\bibitem[Yan et~al\mbox{.}(2023a)]%
        {46}
\bibfield{author}{\bibinfo{person}{Shuanglin Yan}, \bibinfo{person}{Neng Dong},
  \bibinfo{person}{Jun Liu}, \bibinfo{person}{Liyan Zhang}, {and}
  \bibinfo{person}{Jinhui Tang}.} \bibinfo{year}{2023}\natexlab{a}.
\newblock \showarticletitle{Learning comprehensive representations with richer
  self for text-to-image person re-identification}. In
  \bibinfo{booktitle}{\emph{Proceedings of the 31st ACM International
  Conference on Multimedia}}. \bibinfo{pages}{6202--6211}.
\newblock


\bibitem[Yan et~al\mbox{.}(2023b)]%
        {3}
\bibfield{author}{\bibinfo{person}{Shuanglin Yan}, \bibinfo{person}{Neng Dong},
  \bibinfo{person}{Liyan Zhang}, {and} \bibinfo{person}{Jinhui Tang}.}
  \bibinfo{year}{2023}\natexlab{b}.
\newblock \showarticletitle{Clip-driven fine-grained text-image person
  re-identification}.
\newblock \bibinfo{journal}{\emph{IEEE Transactions on Image Processing}}
  \bibinfo{volume}{32} (\bibinfo{year}{2023}), \bibinfo{pages}{6032--6046}.
\newblock


\bibitem[Yan et~al\mbox{.}(2024a)]%
        {47}
\bibfield{author}{\bibinfo{person}{Shuanglin Yan}, \bibinfo{person}{Jun Liu},
  \bibinfo{person}{Neng Dong}, \bibinfo{person}{Liyan Zhang}, {and}
  \bibinfo{person}{Jinhui Tang}.} \bibinfo{year}{2024}\natexlab{a}.
\newblock \showarticletitle{Prototypical Prompting for Text-to-image Person
  Re-identification}. In \bibinfo{booktitle}{\emph{Proceedings of the 32nd ACM
  International Conference on Multimedia}}. \bibinfo{pages}{2331--2340}.
\newblock


\bibitem[Yan et~al\mbox{.}(2024b)]%
        {1}
\bibfield{author}{\bibinfo{person}{Shuanglin Yan}, \bibinfo{person}{Hao Tang},
  \bibinfo{person}{Liyan Zhang}, {and} \bibinfo{person}{Jinhui Tang}.}
  \bibinfo{year}{2024}\natexlab{b}.
\newblock \showarticletitle{Image-specific information suppression and implicit
  local alignment for text-based person search}.
\newblock \bibinfo{journal}{\emph{IEEE Transactions on Neural Networks and
  Learning Systems}} \bibinfo{volume}{35}, \bibinfo{number}{12}
  (\bibinfo{year}{2024}), \bibinfo{pages}{17973--17986}.
\newblock


\bibitem[Yang et~al\mbox{.}(2023)]%
        {36}
\bibfield{author}{\bibinfo{person}{Shuyu Yang}, \bibinfo{person}{Yinan Zhou},
  \bibinfo{person}{Zhedong Zheng}, \bibinfo{person}{Yaxiong Wang},
  \bibinfo{person}{Li Zhu}, {and} \bibinfo{person}{Yujiao Wu}.}
  \bibinfo{year}{2023}\natexlab{}.
\newblock \showarticletitle{Towards unified text-based person retrieval: A
  large-scale multi-attribute and language search benchmark}. In
  \bibinfo{booktitle}{\emph{Proceedings of the 31st ACM International
  Conference on Multimedia}}. \bibinfo{pages}{4492--4501}.
\newblock


\bibitem[Ye et~al\mbox{.}(2021)]%
        {40}
\bibfield{author}{\bibinfo{person}{Mang Ye}, \bibinfo{person}{Jianbing Shen},
  \bibinfo{person}{Gaojie Lin}, \bibinfo{person}{Tao Xiang},
  \bibinfo{person}{Ling Shao}, {and} \bibinfo{person}{Steven~CH Hoi}.}
  \bibinfo{year}{2021}\natexlab{}.
\newblock \showarticletitle{Deep learning for person re-identification: A
  survey and outlook}.
\newblock \bibinfo{journal}{\emph{IEEE Transactions on Pattern Analysis and
  Machine Intelligence}} \bibinfo{volume}{44}, \bibinfo{number}{6}
  (\bibinfo{year}{2021}), \bibinfo{pages}{2872--2893}.
\newblock


\bibitem[Zhang and Lu(2018)]%
        {20}
\bibfield{author}{\bibinfo{person}{Ying Zhang} {and} \bibinfo{person}{Huchuan
  Lu}.} \bibinfo{year}{2018}\natexlab{}.
\newblock \showarticletitle{Deep cross-modal projection learning for image-text
  matching}. In \bibinfo{booktitle}{\emph{Proceedings of the European
  Conference on Computer Vision (ECCV)}}. \bibinfo{pages}{686--701}.
\newblock


\bibitem[Zhao et~al\mbox{.}(2021)]%
        {7}
\bibfield{author}{\bibinfo{person}{Shizhen Zhao}, \bibinfo{person}{Changxin
  Gao}, \bibinfo{person}{Yuanjie Shao}, \bibinfo{person}{Wei-Shi Zheng}, {and}
  \bibinfo{person}{Nong Sang}.} \bibinfo{year}{2021}\natexlab{}.
\newblock \showarticletitle{Weakly supervised text-based person
  re-identification}. In \bibinfo{booktitle}{\emph{Proceedings of the IEEE/CVF
  International Conference on Computer Vision}}. \bibinfo{pages}{11395--11404}.
\newblock


\bibitem[Zheng et~al\mbox{.}(2015)]%
        {41}
\bibfield{author}{\bibinfo{person}{Liang Zheng}, \bibinfo{person}{Liyue Shen},
  \bibinfo{person}{Lu Tian}, \bibinfo{person}{Shengjin Wang},
  \bibinfo{person}{Jingdong Wang}, {and} \bibinfo{person}{Qi Tian}.}
  \bibinfo{year}{2015}\natexlab{}.
\newblock \showarticletitle{Scalable person re-identification: A benchmark}. In
  \bibinfo{booktitle}{\emph{Proceedings of the IEEE International Conference on
  Computer Vision}}. \bibinfo{pages}{1116--1124}.
\newblock


\bibitem[Zheng et~al\mbox{.}(2024)]%
        {8}
\bibfield{author}{\bibinfo{person}{Yanwei Zheng}, \bibinfo{person}{Xinpeng
  Zhao}, \bibinfo{person}{Chuanlin Lan}, \bibinfo{person}{Xiaowei Zhang},
  \bibinfo{person}{Bowen Huang}, \bibinfo{person}{Jibin Yang}, {and}
  \bibinfo{person}{Dongxiao Yu}.} \bibinfo{year}{2024}\natexlab{}.
\newblock \showarticletitle{CPCL: Cross-Modal Prototypical Contrastive Learning
  for Weakly Supervised Text-based Person Re-Identification}.
\newblock \bibinfo{journal}{\emph{arXiv preprint arXiv:2401.10011}}
  (\bibinfo{year}{2024}).
\newblock


\bibitem[Zheng et~al\mbox{.}(2020)]%
        {22}
\bibfield{author}{\bibinfo{person}{Zhedong Zheng}, \bibinfo{person}{Liang
  Zheng}, \bibinfo{person}{Michael Garrett}, \bibinfo{person}{Yi Yang},
  \bibinfo{person}{Mingliang Xu}, {and} \bibinfo{person}{Yi-Dong Shen}.}
  \bibinfo{year}{2020}\natexlab{}.
\newblock \showarticletitle{Dual-path convolutional image-text embeddings with
  instance loss}.
\newblock \bibinfo{journal}{\emph{ACM Transactions on Multimedia Computing,
  Communications, and Applications (TOMM)}} \bibinfo{volume}{16},
  \bibinfo{number}{2} (\bibinfo{year}{2020}), \bibinfo{pages}{1--23}.
\newblock


\bibitem[Zhu et~al\mbox{.}(2021)]%
        {14}
\bibfield{author}{\bibinfo{person}{Aichun Zhu}, \bibinfo{person}{Zijie Wang},
  \bibinfo{person}{Yifeng Li}, \bibinfo{person}{Xili Wan},
  \bibinfo{person}{Jing Jin}, \bibinfo{person}{Tian Wang},
  \bibinfo{person}{Fangqiang Hu}, {and} \bibinfo{person}{Gang Hua}.}
  \bibinfo{year}{2021}\natexlab{}.
\newblock \showarticletitle{Dssl: Deep surroundings-person separation learning
  for text-based person retrieval}. In \bibinfo{booktitle}{\emph{Proceedings of
  the 29th ACM International Conference on Multimedia}}.
  \bibinfo{pages}{209--217}.
\newblock


\end{thebibliography}


\end{document}